\documentclass{article}
\usepackage[letterpaper, margin=2cm]{geometry}
\usepackage[bookmarks=false,pdfstartview={FitH}]{hyperref}
\usepackage{breakurl}
\usepackage{graphicx}
\usepackage{paralist}
\usepackage{amsmath}
\usepackage{subfigure}
\usepackage{multirow}
\usepackage{fancyhdr}

\title{Lesion Border Detection in Dermoscopy Images\\Using Ensembles of Thresholding Methods}

\author{%
        M. Emre Celebi\\
        Dept.\ of Computer Science\\Louisiana State Univ., Shreveport, LA, USA\\
        \href{mailto:ecelebi@lsus.edu}{ecelebi@lsus.edu}\\
\and
        Quan Wen\\
        School of Computer Science and Engineering\\Univ.\ of Electronic Science and Technology of China, Chengdu, P.R.\ China\\
        \href{mailto:quanwen@uestc.edu.cn}{quanwen@uestc.edu.cn}\\
\and
        Sae Hwang\\
        Dept.\ of Computer Science\\Univ.\ of Illinois, Springfield, IL, USA\\
        \href{mailto:shwan2@uis.edu}{shwan2@uis.edu}\\
\and
        Hitoshi Iyatomi\\
        Dept.\ of Applied Informatics\\Hosei Univ., Tokyo, Japan\\
        \href{mailto:iyatomi@hosei.ac.jp}{iyatomi@hosei.ac.jp}\\
\and
        Gerald Schaefer\\
        Dept.\ of Computer Science\\Loughborough Univ., Loughborough, UK\\
        \href{mailto:gerald.schaefer@ieee.org}{gerald.schaefer@ieee.org}
}

\DeclareMathOperator{\I}{I}
\DeclareMathOperator{\U}{U}

\begin{document}

\maketitle
\begin{abstract}

Dermoscopy is one of the major imaging modalities used in the diagnosis of melanoma and other pigmented skin lesions. Due to the difficulty and subjectivity of human interpretation, automated analysis of dermoscopy images has become an important research area. Border detection is often the first step in this analysis. In many cases, the lesion can be roughly separated from the background skin using a thresholding method applied to the blue channel. However, no single thresholding method appears to be robust enough to successfully handle the wide variety of dermoscopy images encountered in clinical practice. In this paper, we present an automated method for detecting lesion borders in dermoscopy images using ensembles of thresholding methods. Experiments on a difficult set of 90 images demonstrate that the proposed method is robust, fast, and accurate when compared to nine state-of-the-art methods.

\end{abstract}

\section{Introduction}

Invasive and in-situ malignant melanoma together comprise one of the most rapidly increasing cancers in the world. Invasive melanoma alone has an estimated incidence of 70,230 and an estimated total of 8,790 deaths in the United States in 2011 \cite{Siegel11}. Early diagnosis is particularly important since melanoma can be cured with a simple excision if detected early.
\par
Dermoscopy has become one of the most important tools in the diagnosis of melanoma and other pigmented skin lesions. This non-invasive skin imaging technique involves optical magnification and either liquid immersion and low angle-of-incidence lighting or cross-polarized lighting, making subsurface structures more easily visible when compared to conventional clinical images \cite{Argenziano02}. Dermoscopy allows the identification of dozens of morphological features such as pigment networks, dots/globules, streaks, blue-white areas, and blotches \cite{Menzies09}. This reduces screening errors and provides greater differentiation between difficult lesions such as pigmented Spitz nevi and small, clinically equivocal lesions \cite{Steiner93}. However, it has also been demonstrated that dermoscopy may actually lower the diagnostic accuracy in the hands of inexperienced dermatologists \cite{Binder95}. Therefore, in order to minimize the diagnostic errors that result from the difficulty and subjectivity of visual interpretation, the development of computerized image analysis techniques is of paramount importance \cite{Celebi07a}.
\par
Automated border detection is often the first step in the automated or semi-automated analysis of dermoscopy images \cite{Celebi09}. It is crucial for image analysis for two main reasons. First, the border structure provides important information for accurate diagnosis, as many clinical features, such as asymmetry, border irregularity, and abrupt border cutoff, are calculated directly from the border. Second, extraction of other important clinical features such as atypical pigment networks, globules, and blue-white areas, critically depends on the accuracy of border detection. Automated border detection is a challenging task due to several reasons:
\begin{inparaenum}[(i)]
\item low contrast between the lesion and the surrounding skin,
\item irregular and fuzzy lesion borders,
\item artifacts and intrinsic cutaneous features such as black frames, skin lines, blood vessels, hairs, and air bubbles,
\item variegated coloring inside the lesion, and
\item fragmentation due to various reasons such as scar-like depigmentation.
\end{inparaenum}
\par
Numerous methods have been developed for border detection in dermoscopy images \cite{Celebi09}. Recent approaches include thresholding \cite{Iyatomi08, Garnavi11}, k-means clustering \cite{Zhou08}, fuzzy c-means clustering \cite{Schmid99, Zhou09}, density-based clustering \cite{Mete11}, meanshift clustering \cite{Melli06}, gradient vector flow snakes \cite{Erkol05, Zhou11, Abbas12b}, color quantization followed by spatial segmentation \cite{Celebi07b}, statistical region merging \cite{Celebi08}, watershed transformation \cite{Wang11}, dynamic programming \cite{Abbas11, Abbas12a}, and supervised learning \cite{Schaefer11, Wighton11}.
\par
In this paper, we present a fast and accurate method for detecting lesion borders in dermoscopy images. The method involves the fusion several thresholding methods followed by various simple postprocessing steps. The rest of the paper is organized as follows. Section \ref{sec_fusion} describes the threshold fusion method and the postprocessing steps. Section \ref{sec_exp} presents the experimental results. Finally, Section \ref{sec_conc} gives the conclusions.

\section{Threshold Fusion}
\label{sec_fusion}

\begin{figure}[!ht]
\centering
 \subfigure[Original image]{\label{threshold_a}\includegraphics[width=0.36\columnwidth]{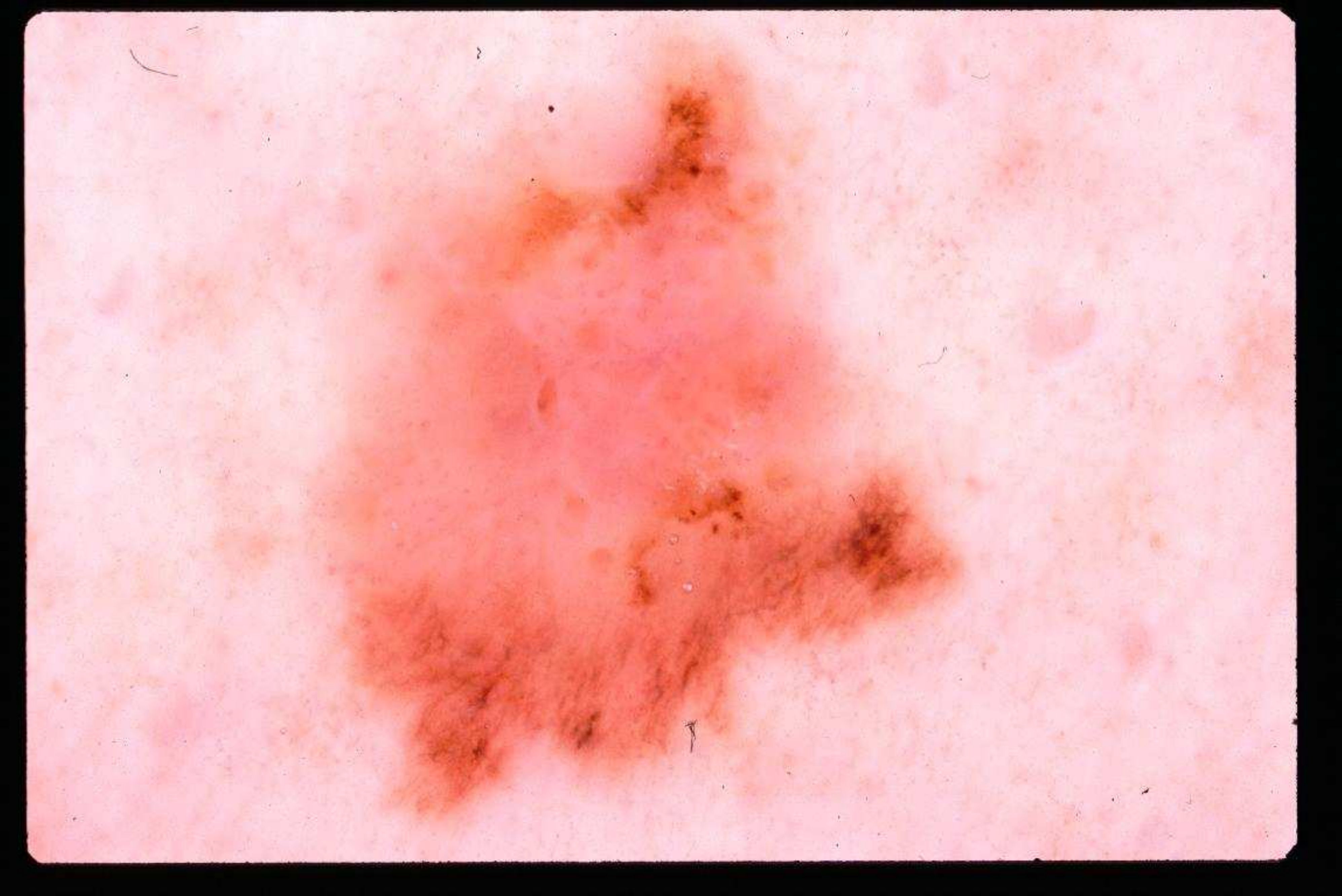}}
 \hspace{.05in}
 \subfigure[Blue channel]{\label{threshold_b}\includegraphics[width=0.36\columnwidth]{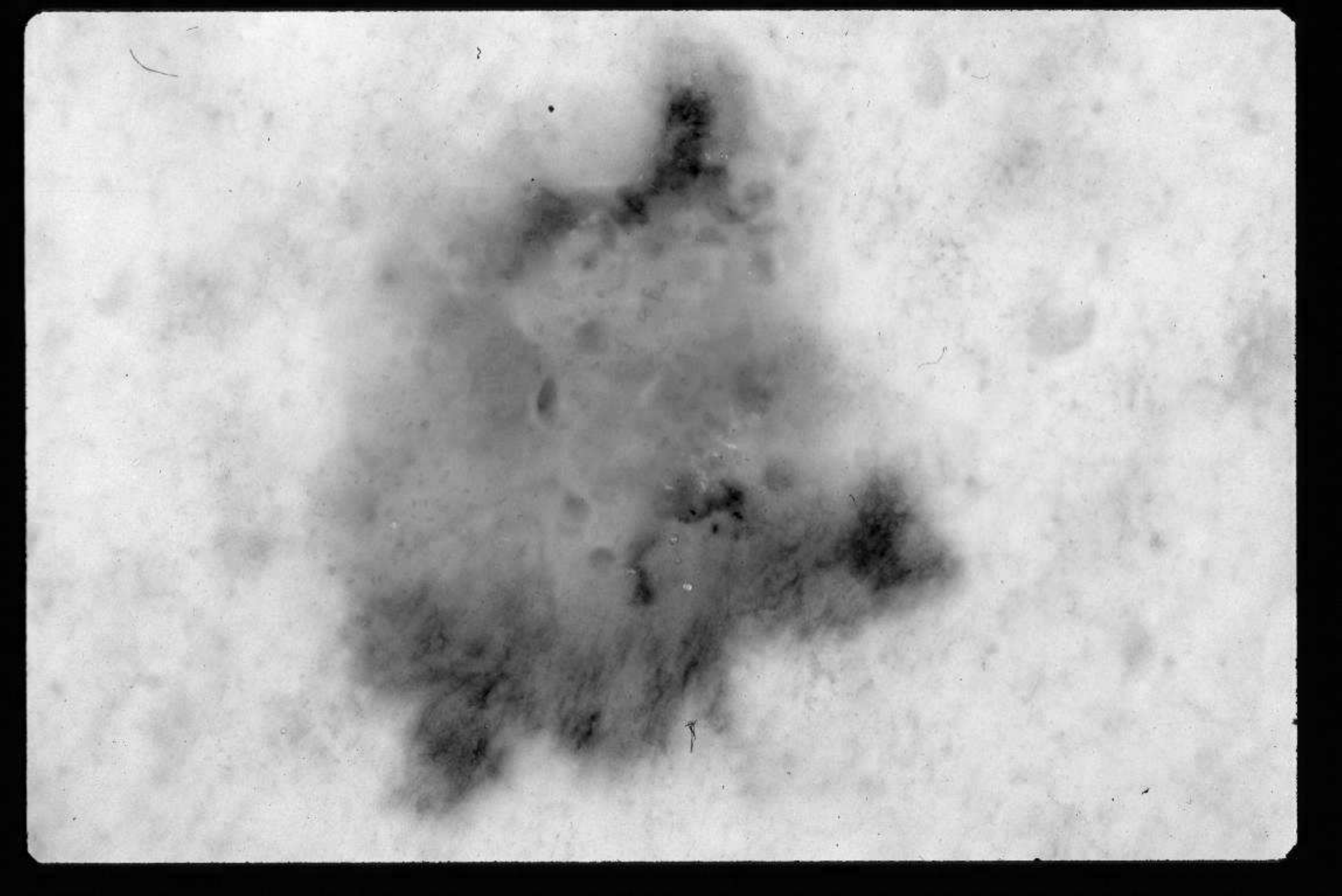}}
 \subfigure[Huang \& Wang's method \cite{Huang95} ($T = 183$)]{\label{threshold_c}\includegraphics[width=0.36\columnwidth]{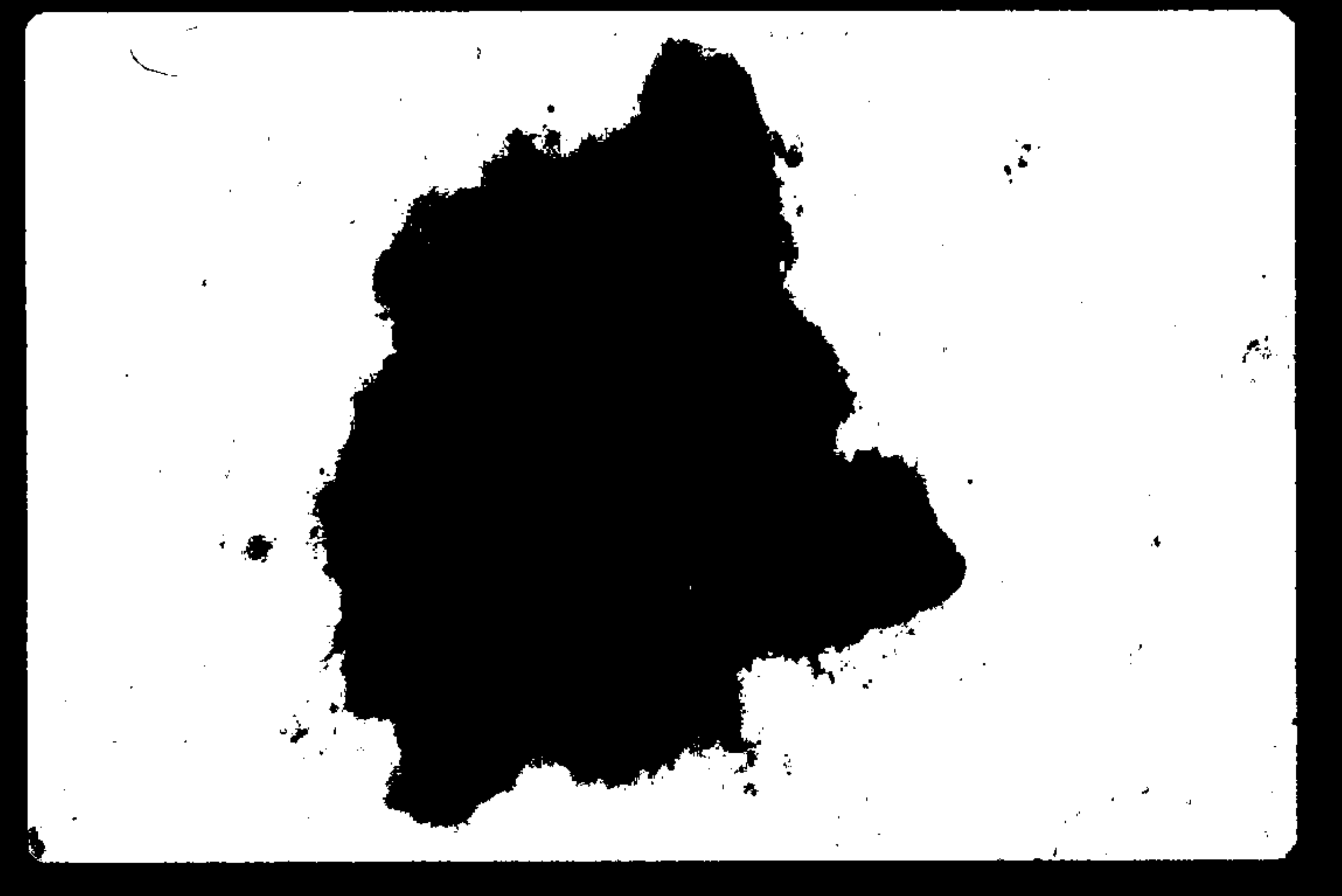}}
 \hspace{.05in}
 \subfigure[Kapur \emph{et al.}'s method \cite{Kapur85} ($T = 178$)]{\label{threshold_d}\includegraphics[width=0.36\columnwidth]{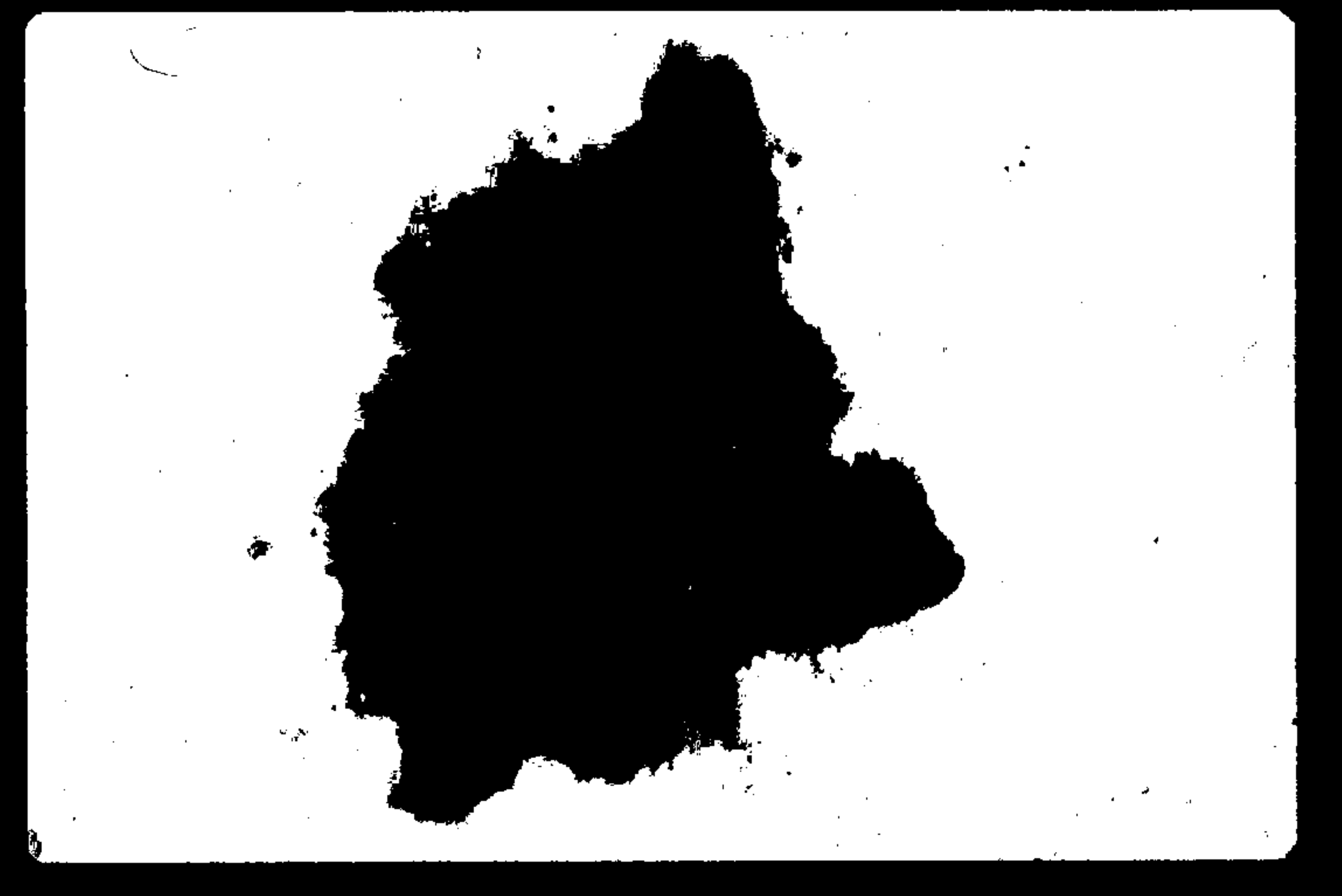}}
 \subfigure[Kittler \& Illingworth's method \cite{Kittler86} ($T = 192$)]{\label{threshold_e}\includegraphics[width=0.36\columnwidth]{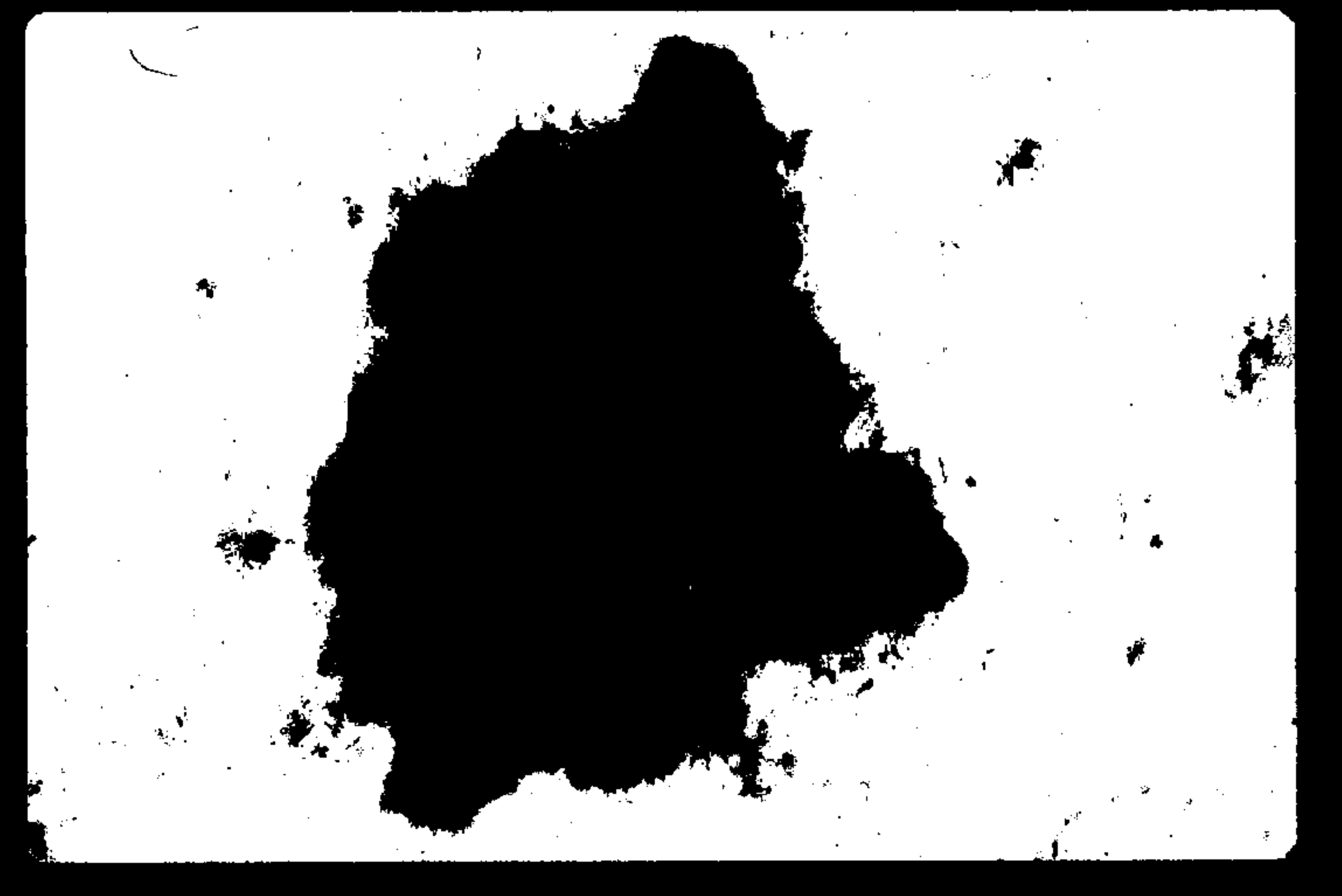}}
 \hspace{.05in}
 \subfigure[Otsu's method \cite{Otsu79} ($T = 137$)]{\label{threshold_f}\includegraphics[width=0.36\columnwidth]{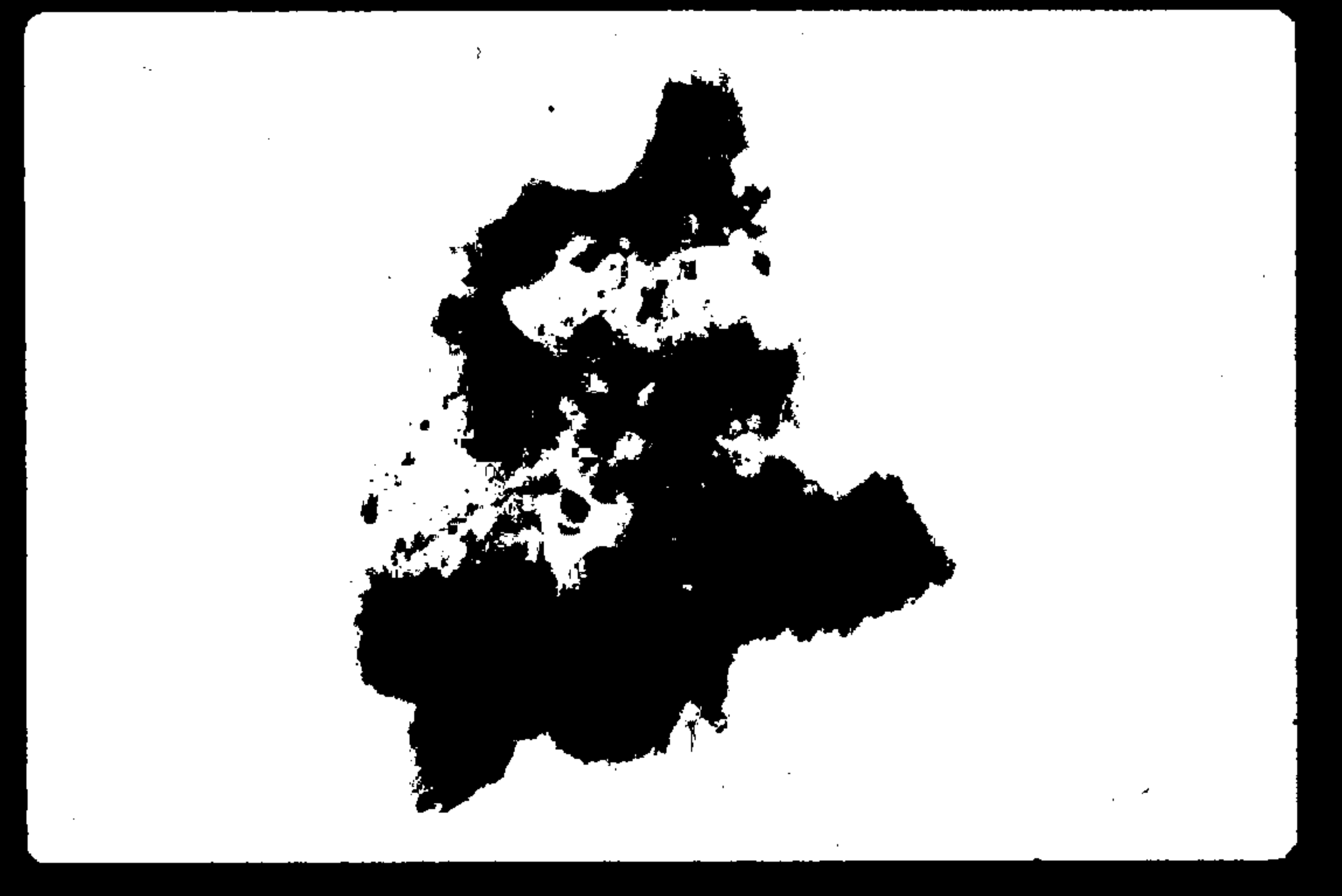}}
 \caption{Comparison of various thresholding methods (\emph{T}: threshold)}
 \label{fig_thresh_comp}
\end{figure}

In many dermoscopic images, the lesion can be roughly separated from the background skin using a thresholding method applied to the blue channel \cite{Celebi09}. While there are a number of thresholding methods that perform well in general, the effectiveness of a method strongly depends on the statistical characteristics of the image \cite{Melgani06}. Fig.\ \ref{fig_thresh_comp} illustrates this phenomenon. Here, method \ref{threshold_e} performs quite well. In contrast, methods \ref{threshold_c}, \ref{threshold_d}, and \ref{threshold_f} underestimate the optimal threshold. Although method \ref{threshold_f} is the most popular thresholding method in the literature \cite{Sezgin04}, for this particular image, it performs the worst.

A possible approach to overcome this problem is to fuse the results provided by an ensemble of thresholding methods. In this way, it is  possible to exploit the peculiarities of the participating thresholding methods synergistically, thus arriving at more robust final decisions than is possible with a single thresholding method. It should be noted that the goal of the fusion is not to outperform the individual thresholding methods, but to obtain accuracies comparable to that of the best thresholding method independently of the image characteristics. In this study, we adopted the threshold fusion method proposed by Melgani \cite{Melgani06}, which we describe briefly in the following.
\par
Let $X=\{x_{mn}: m=0,1,\ldots,M-1,\ n=0,1,\ldots$ $,N-1\}$ be the original scalar $M \times N$ image with $L$ possible gray levels $\left(x_{mn}\in\left\{0,1,\ldots,L-1\right\}\right)$ and $Y=\left\{y_{mn}: m=0,1,\ldots,M-1,\ n=0,1,\ldots,N-1\right\}$ be the binary output of the threshold fusion. Consider an ensemble of $P$ thresholding methods. Let $T_i$ and $A_i \ (i=1,2,\ldots,P)$ be the threshold value and the output binary image associated with the $i$-th method of the ensemble, respectively. Within a Markov Random Field (MRF) framework, the fusion problem can be formulated as an energy minimization task. Accordingly, the local energy function $\U_{mn}$ to be minimized for the pixel $(m,n)$ can be written as follows:
\begin{equation}
\label{energy_function}
\U_{mn} = \beta_{SP} \cdot \U_{SP}\left[y_{mn},Y^S(m,n)\right] + \sum_{i=1}^P { \beta_i \cdot \U_{II}\left[y_{mn},A^S_{i}(m,n)\right] }
\end{equation}
where $S$ is a predefined neighborhood system associated with pixel $(m,n)$, $\U_{SP}(\cdot)$ and $\U_{II}(\cdot)$ refer  to the spatial and inter-image energy functions, respectively, whereas $\beta_{SP}$ and $\beta_i \ (i=1,2,\ldots,P)$ represent the spatial and inter-image parameters, respectively. The spatial energy function can be expressed as:
\begin{equation}
\U_{SP}\left[y_{mn},Y^S(m,n)\right] = - \sum_{ y_{pq} \in Y^S(m,n) } \I\left(y_{mn},y_{pq}\right)
\end{equation}
where $\I(.,.)$ is the indicator function defined as:
\begin{equation}
\I{\left(y_{mn},y_{pq}\right)}= \left\{
                 \begin{array}
                 {r@{\quad}l}
                 1 & \mbox{if} \ y_{mn}=y_{pq} \\
                 0 & \mbox{otherwise}
                 \end{array}
          \right.
\end{equation}

The inter-image energy function is defined as:
\begin{equation}
\U_{II}\left[y_{mn},A^S(m,n)\right] = - \sum_{ A_i(p,q) \in A^S_{i}(m,n) } \alpha^i(x_{pq}) \cdot \I\left[y_{mn},A_i(p,q)\right]
\end{equation}
where $\alpha^i(x_{mn}) = 1 - \exp \left( -\gamma \left| x_{mn} - T_i \right| \right)$ is a weight function.

This function controls the effect of unreliable decisions at the pixel level that can be incurred by the thresholding methods. At the global (image) level,  decisions are weighed by the inter-image parameters $\beta_i \ (i=1,2,\ldots,P)$, which are computed as follows:
$\beta_i = \exp \left( -\gamma \left| \bar{T} - T_i \right| \right)$
where $\bar{T}$ is the average threshold value, i.e.\ $\bar{T} = (1/P) \sum_{i=1}^P T_i$.

The MRF fusion strategy proposed in \cite{Melgani06} is as follows:

\begin{enumerate}
  \item Apply each thresholding method of the ensemble to the image $X$ to generate the set of thresholded images $A_i \ (i=1,2,\ldots,P)$.
  \item Initialize $Y$ by minimizing for each pixel $(m,n)$ the local energy function $\U_{mn}$ defined in Eq.\ \eqref{energy_function} without the spatial energy term, i.e.\ by setting $\beta_{SP}=0$.
  \item Update $Y$ by minimizing for each pixel $(m,n)$ the local energy function $\U_{mn}$ defined in Eq.\ \eqref{energy_function} including the spatial energy term, i.e.\ by setting $\beta_{SP} \neq 0$.
  \item Repeat step (3) $K_{max}$ times or until the number of different labels in $Y$ computed over the last two iterations becomes very small.
\end{enumerate}

In our preliminary experiments, we observed that, besides being computationally demanding, the iterative part (step 3) of the fusion method makes only marginal contribution to the quality of the results. Therefore, in this study, we considered only the first two steps. The $\gamma$ parameter was set to the recommended value of $0.1$ \cite{Melgani06}. For computational reasons, $\alpha$ and $\beta$ values were precalculated and the neighborhood system $S$ was chosen as a $3 \times 3$ square. We considered four popular thresholding methods to construct the ensemble: Huang \& Wang's fuzzy similarity method \cite{Huang95}, Kapur \emph{et al.}'s maximum entropy method \cite{Kapur85}, Kittler \& Illingworth's minimum error thresholding method \cite{Kittler86}, and Otsu's clustering based method \cite{Otsu79}.
\par
\begin{figure}[!ht]
\centering
\includegraphics[width=0.4\columnwidth]{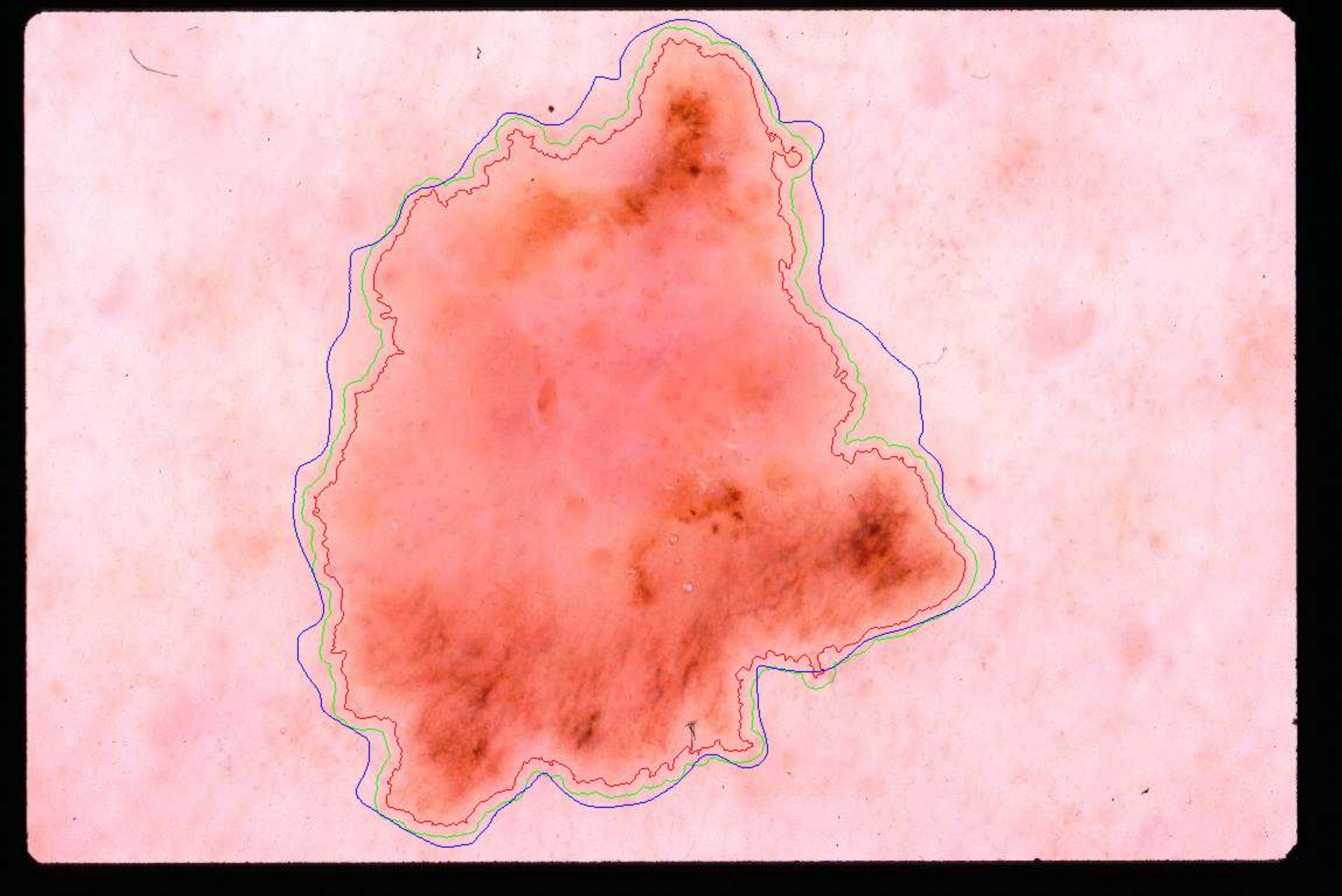}
\caption{\label{fig_fusion} Fusion result for Fig.\ \ref{threshold_a}}
\end{figure}

After performing the threshold fusion on the blue channel, the final border was obtained by filling the binary fusion output and removing all but the largest (4-)connected component in it. Fig.\ \ref{fig_fusion} shows the fusion result for the image given in Fig.\ \ref{threshold_a}. Here the fusion output is delineated in red, whereas the manual border (see Section \ref{sec_exp}) is delineated in blue. It can be seen that the former is mostly contained inside the latter. This was observed in many cases because the automated methods tend to find the sharpest pigment change, whereas the dermatologists choose the outmost detectable pigment \cite{Celebi09}. The discrepancy between the two borders can be reduced by expanding the fusion output using morphological dilation. A circular structuring element with a radius of
$R = \left\lfloor {kD/512} \right\rfloor$, where $\lfloor . \rfloor$ and $D$ denote the floor operation and diameter of the lesion object in the fusion output,  respectively and $k$ is a scaling factor, which in this study was set to $k=7$. It can be seen that after the expansion, the fusion output (delineated in green) is significantly closer to the manual border, which is also reflected by a reduction in the XOR error (see Section \ref{sec_exp}) from $17.14\%$ to $8.79\%$.

\section{Experimental Results and Discussion}
\label{sec_exp}

The proposed border detection method was tested on a set of 90 dermoscopy images (23 invasive malignant melanoma and 67 benign) obtained from the EDRA Interactive Atlas of Dermoscopy \cite{Argenziano02}, and three private dermatology practices \cite{Celebi08}. An experienced dermatologist determined the manual borders by selecting a number of points on the lesion border, which were then connected by a second-order B-spline. The presented fusion method was compared to nine recent methods using the XOR measure \cite{Hance96} given by
$
\varepsilon={{\mbox{Area}(AB \oplus MB)} \mathord{\left/
 {\vphantom {{\mbox{Area}(AB \oplus MB)} {\mbox{Area}(MB)}}} \right.
 \kern-\nulldelimiterspace} {\mbox{Area}(MB)}}
$,
where \emph{AB} is the binary output of an automated method, \emph{MB} is the binary border drawn by the dermatologist, $\oplus$ is the exclusive-OR operation that determines the pixels for which the \emph{AB} and \emph{MB} disagree, and $\mbox{Area}(I)$ denotes the number of pixels in the binary image $I$.
\par
Table \ref{tab_ensemble} gives the mean ($\mu$) and standard deviation ($\sigma$) percent XOR errors for the ten automated methods. It can be seen that the presented fusion method is significantly more accurate when compared to the other methods. Furthermore, our method is more stable as evidenced by its low standard deviation.

\begin{table}
\centering
\caption{ \label{tab_ensemble} Error statistics for the border detection methods }
\begin{tabular}{l|rr|rr|rr}
\hline
 & \multicolumn{2}{c|}{Benign} & \multicolumn{2}{c|}{Melanoma} & \multicolumn{2}{c}{All}\\
Method & \multicolumn{1}{c}{$\mu$} & \multicolumn{1}{c|}{$\sigma$} & \multicolumn{1}{c}{$\mu$} & \multicolumn{1}{c|}{$\sigma$} & \multicolumn{1}{c}{$\mu$} & \multicolumn{1}{c}{$\sigma$}\\
\hline
\cite{Schmid99} & 22.99 & 12.61 & 28.31 & 15.25 & 24.35 & 13.45\\
\cite{Erkol05} & 13.69 & 5.59 & 19.34 & 9.33 & 15.13 & 7.13\\
\cite{Iyatomi08} & 10.51 & 4.73 & 11.85 & 6.00 & 10.86 & 5.08\\
\cite{Melli06} & 11.53 & 9.74 & 13.29 & 7.42 & 11.98 & 9.19\\
\cite{Celebi07b} & 10.83 & 6.36 & 13.75 & 7.59 & 11.58 & 6.77\\
\cite{Celebi08} & 11.38 & 6.23 & 10.29 & 5.84 & 11.11 & 6.12\\
\cite{Zhou08} & 21.56 & 25.19 & 23.51 & 16.06 & 22.06 & 23.13\\
\cite{Garnavi11} & 12.95 & 6.17 & 16.93 & 7.16 & 13.96 & 6.63\\
\cite{Schaefer11} & 10.07 & 4.34 & 18.17 & 26.96 & 12.14 & 14.36\\
\emph{Fusion} & \textbf{8.36} & \textbf{4.33} & \textbf{8.17} & \textbf{3.13} & \textbf{8.31} & \textbf{4.06}\\
\hline
\end{tabular}
\end{table}

Table \ref{tab_ind} shows the statistics for the individual thresholding methods. Note that, the outputs of these methods were postprocessed as described in Section \ref{sec_fusion}. It can be seen that the individual methods obtain significantly higher mean errors when compared to the fusion method. This is because, as explained in Section \ref{sec_fusion}, the individual methods are more prone to catastrophic failures when given pathological input images. The high standard deviation values also support this explanation. Only the performance of Otsu's method is close to the performance of the fusion. However, as mentioned earlier, the goal of fusion is not to outperform the individual thresholding methods, but to obtain results comparable to that of the best thresholding method independently of the image characteristics.

\begin{table}
\centering
\caption{ \label{tab_ind} Error statistics for the individual thresholding methods}
\begin{tabular}{l|rr|rr|rr}
\hline
 & \multicolumn{2}{c|}{Benign} & \multicolumn{2}{c|}{Melanoma} & \multicolumn{2}{c}{All}\\
Method & \multicolumn{1}{c}{$\mu$} & \multicolumn{1}{c|}{$\sigma$} & \multicolumn{1}{c}{$\mu$} & \multicolumn{1}{c|}{$\sigma$} & \multicolumn{1}{c}{$\mu$} & \multicolumn{1}{c}{$\sigma$}\\
\hline
Huang \& Wang & 9.01 & 5.43 & 15.65 & 29.82 & 10.70 & 15.82\\
Kapur \emph{et al.} & 20.99 & 18.20 & 14.88 & 14.21 & 19.43 & 17.40\\
Kittler \& Illingworth & 10.08 & 8.08 & 21.34 & 38.17 & 12.95 & 20.81\\
Otsu & \textbf{8.66} & \textbf{4.93} & \textbf{10.64} & \textbf{5.81} & \textbf{9.16} & \textbf{5.21}\\
\hline
\end{tabular}
\end{table}

\begin{figure}[!ht]
\centering
 \subfigure[Benign ($\varepsilon = 2.05\%$)]{\label{result_a}\includegraphics[width=0.36\columnwidth]{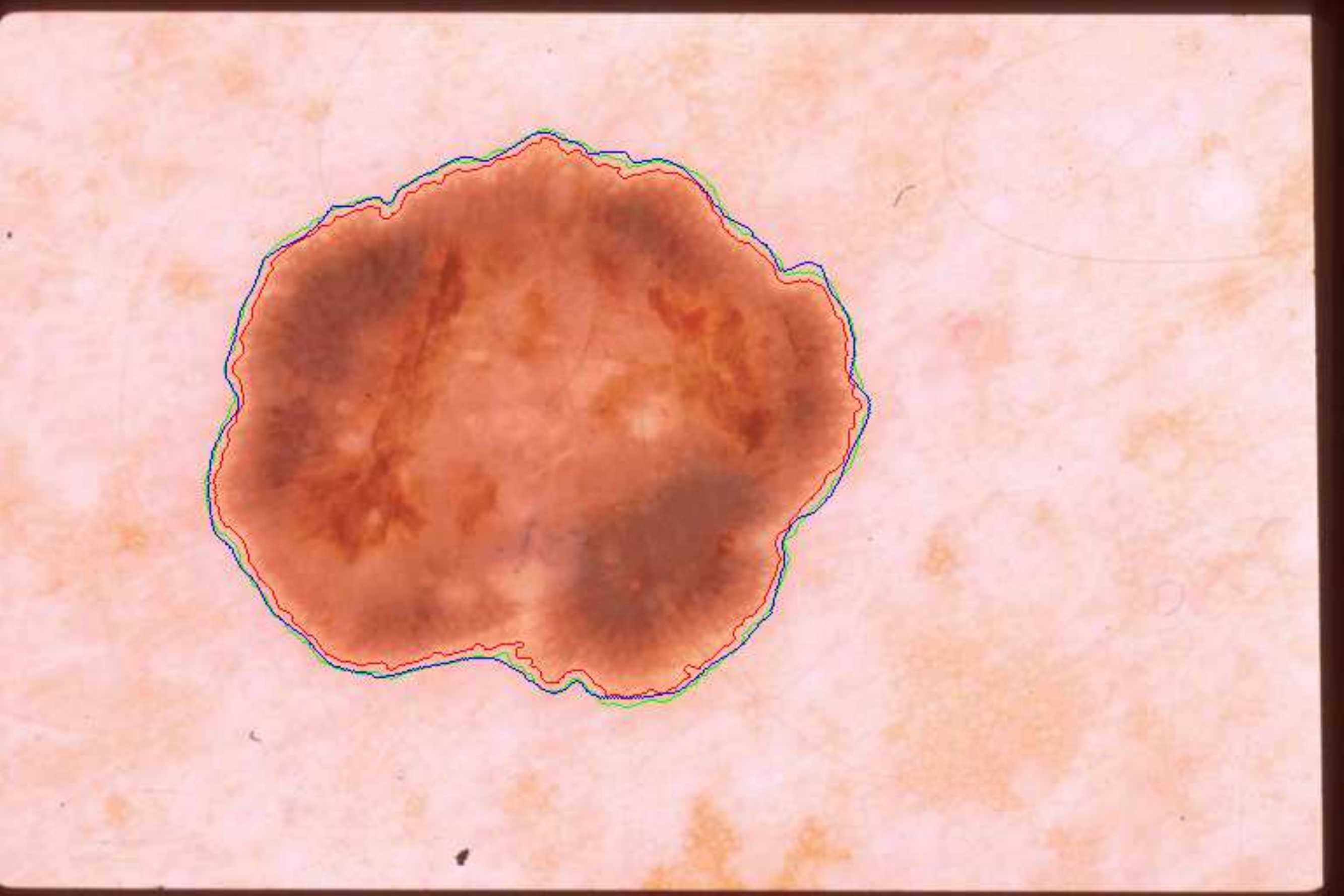}}
 \hspace{.05in}
 \subfigure[Melanoma ($\varepsilon = 4.05\%$)]{\label{result_b}\includegraphics[width=0.36\columnwidth]{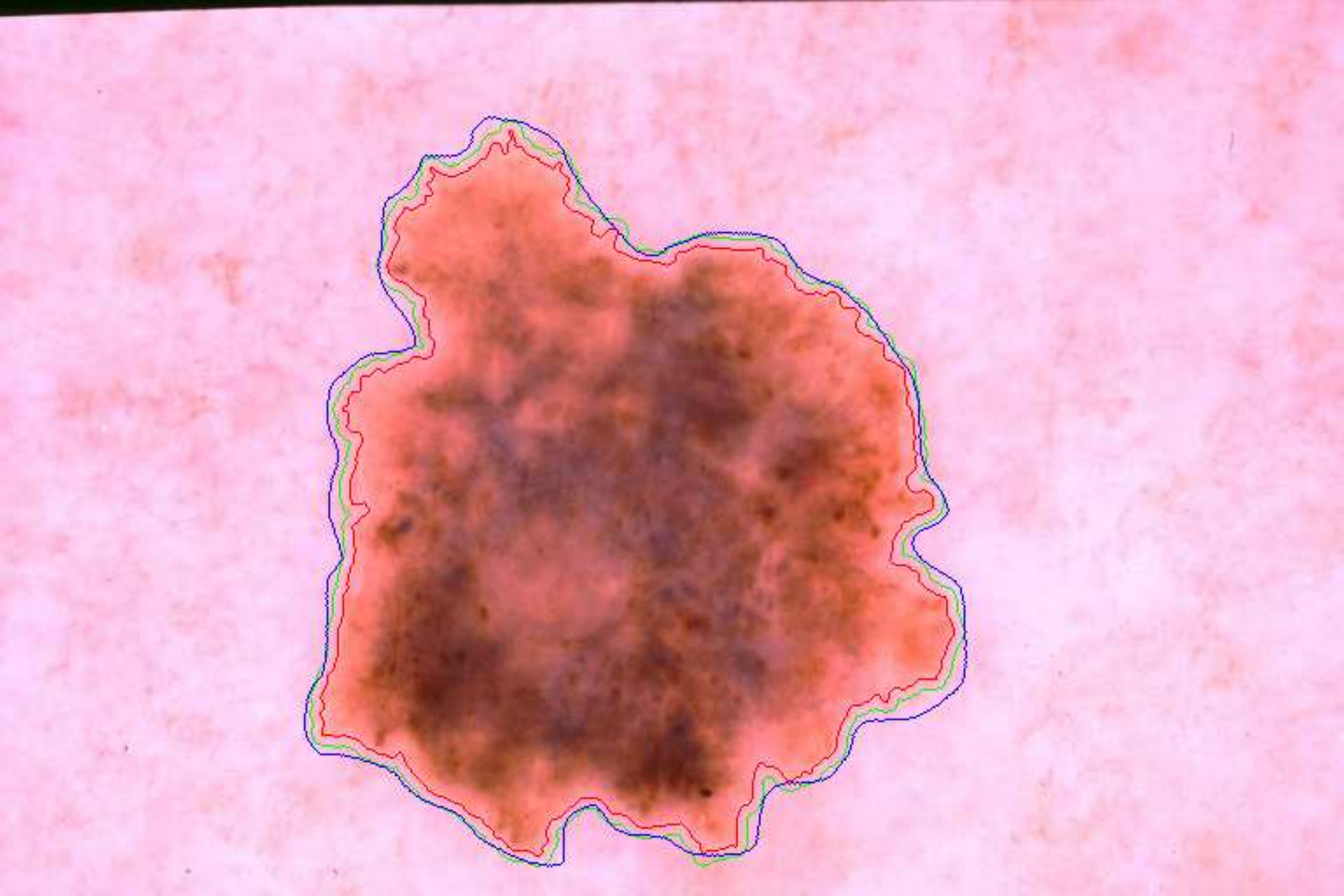}}
 \\
 \subfigure[Melanoma ($\varepsilon = 5.99\%$)]{\label{result_c}\includegraphics[width=0.36\columnwidth]{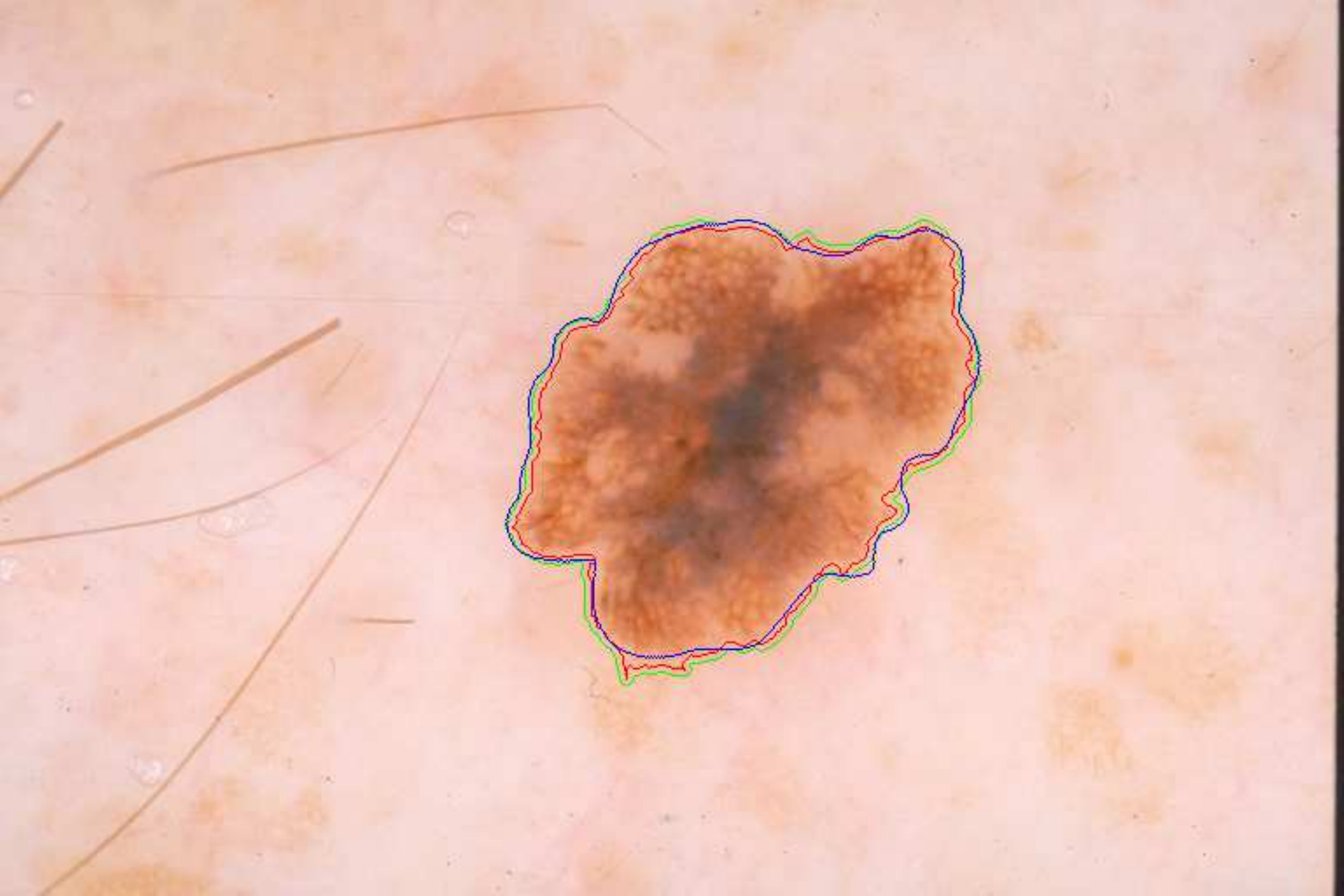}}
 \hspace{.05in}
 \subfigure[Melanoma ($\varepsilon = 8.06\%$)]{\label{result_d}\includegraphics[width=0.36\columnwidth]{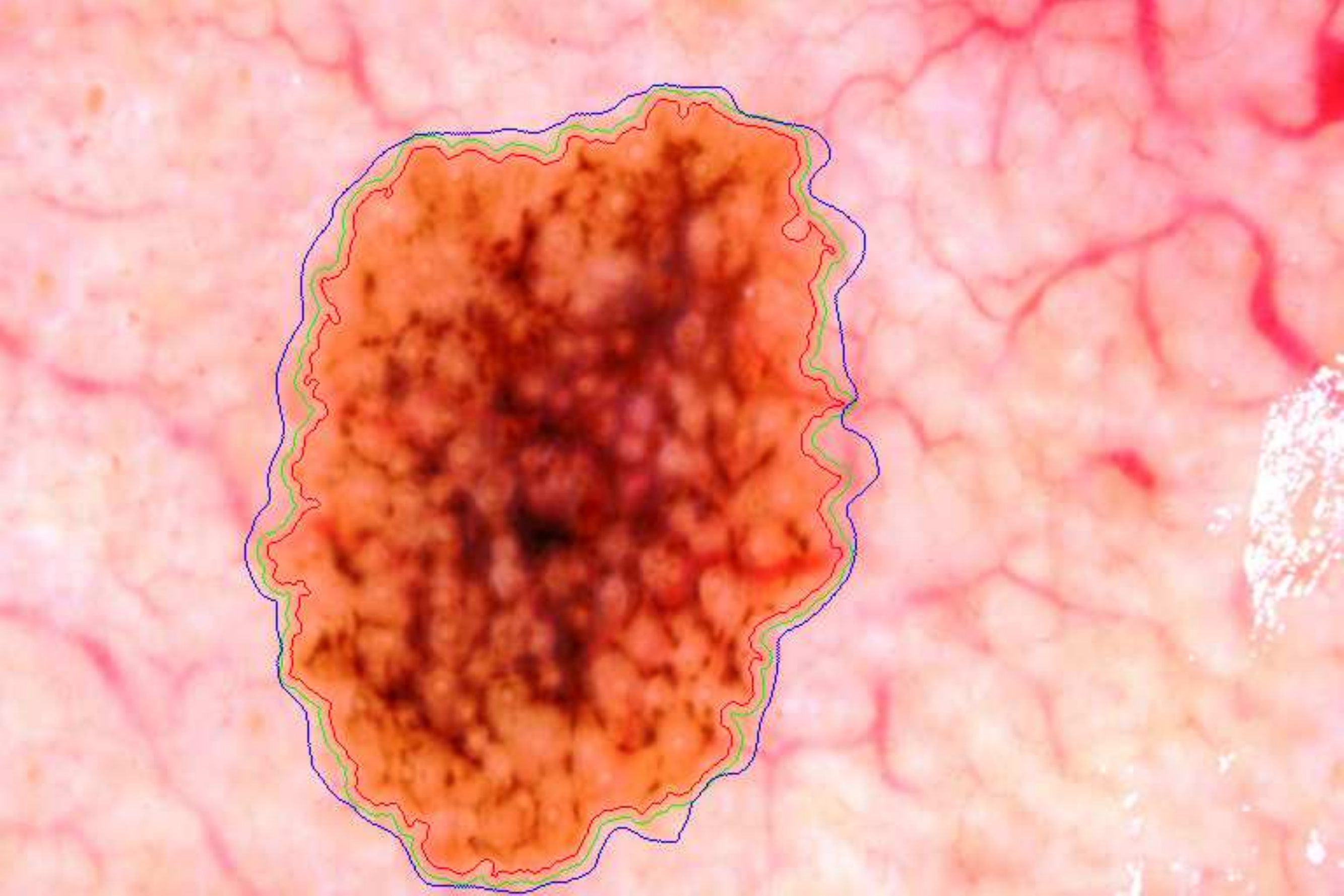}}
 \\
 \subfigure[Benign ($\varepsilon = 10.07\%$)]{\label{result_e}\includegraphics[width=0.36\columnwidth]{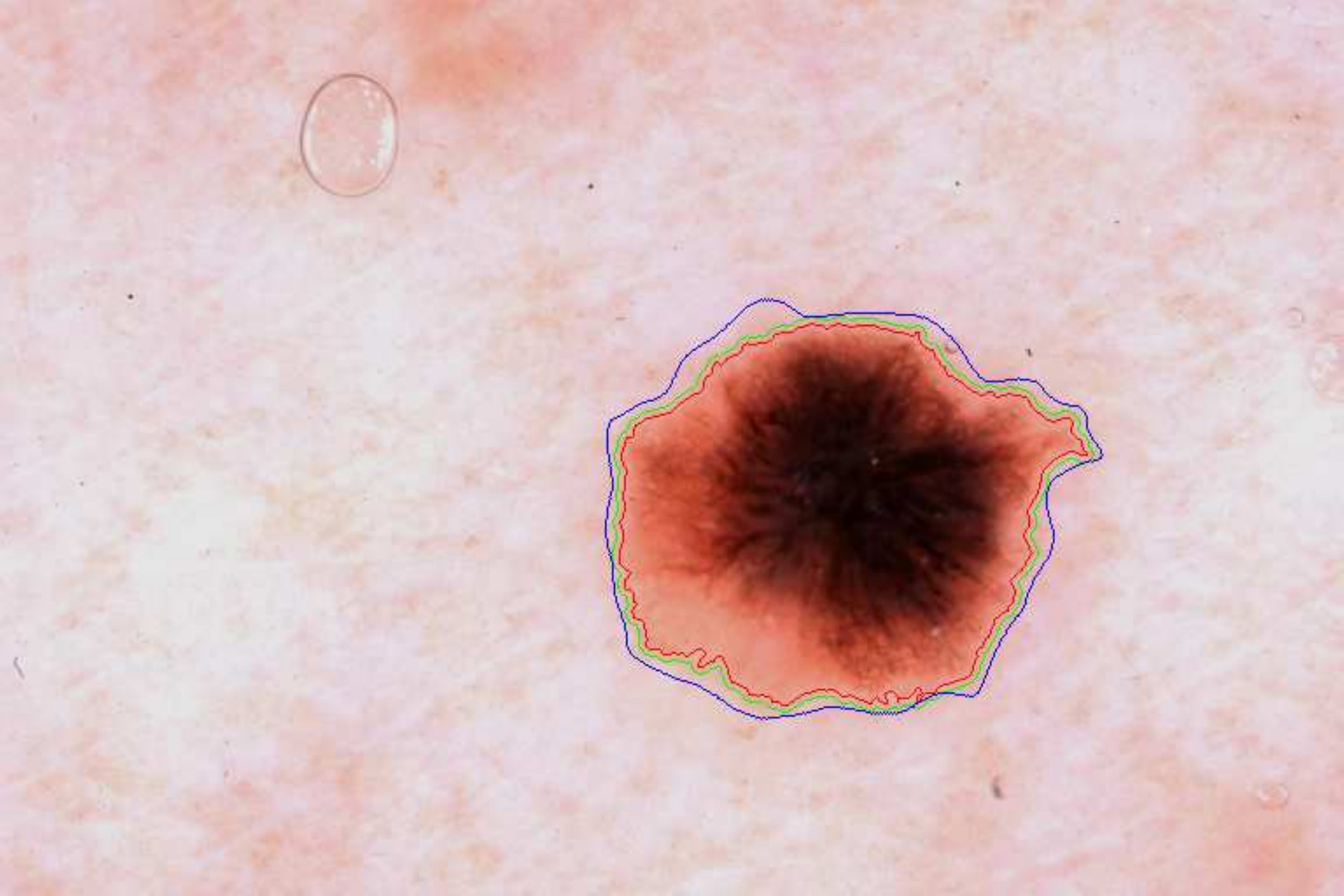}}
 \hspace{.05in}
 \subfigure[Benign ($\varepsilon = 12.14\%$)]{\label{result_f}\includegraphics[width=0.36\columnwidth]{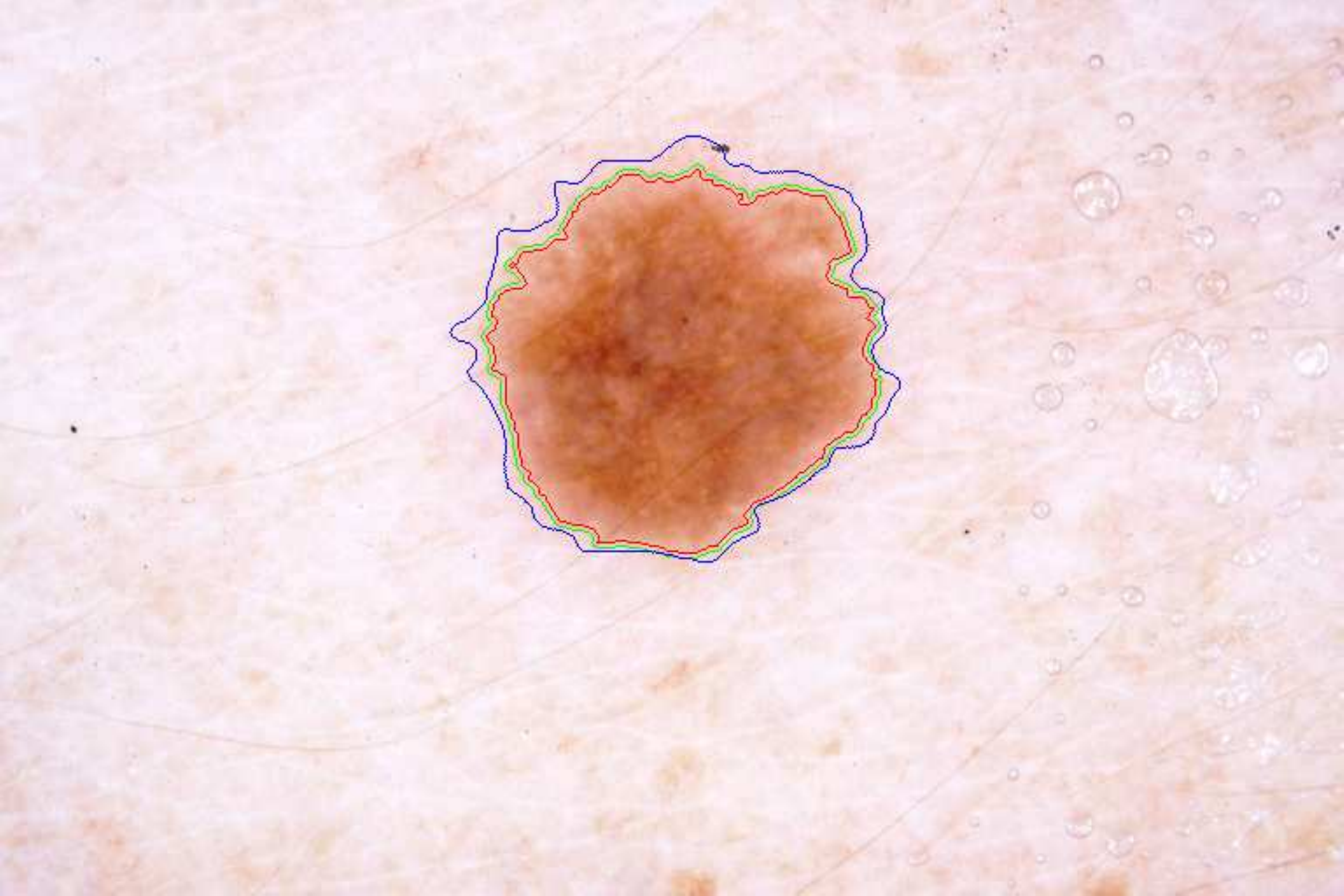}}
 \\
 \subfigure[Benign ($\varepsilon = 14.21\%$)]{\label{result_g}\includegraphics[width=0.36\columnwidth]{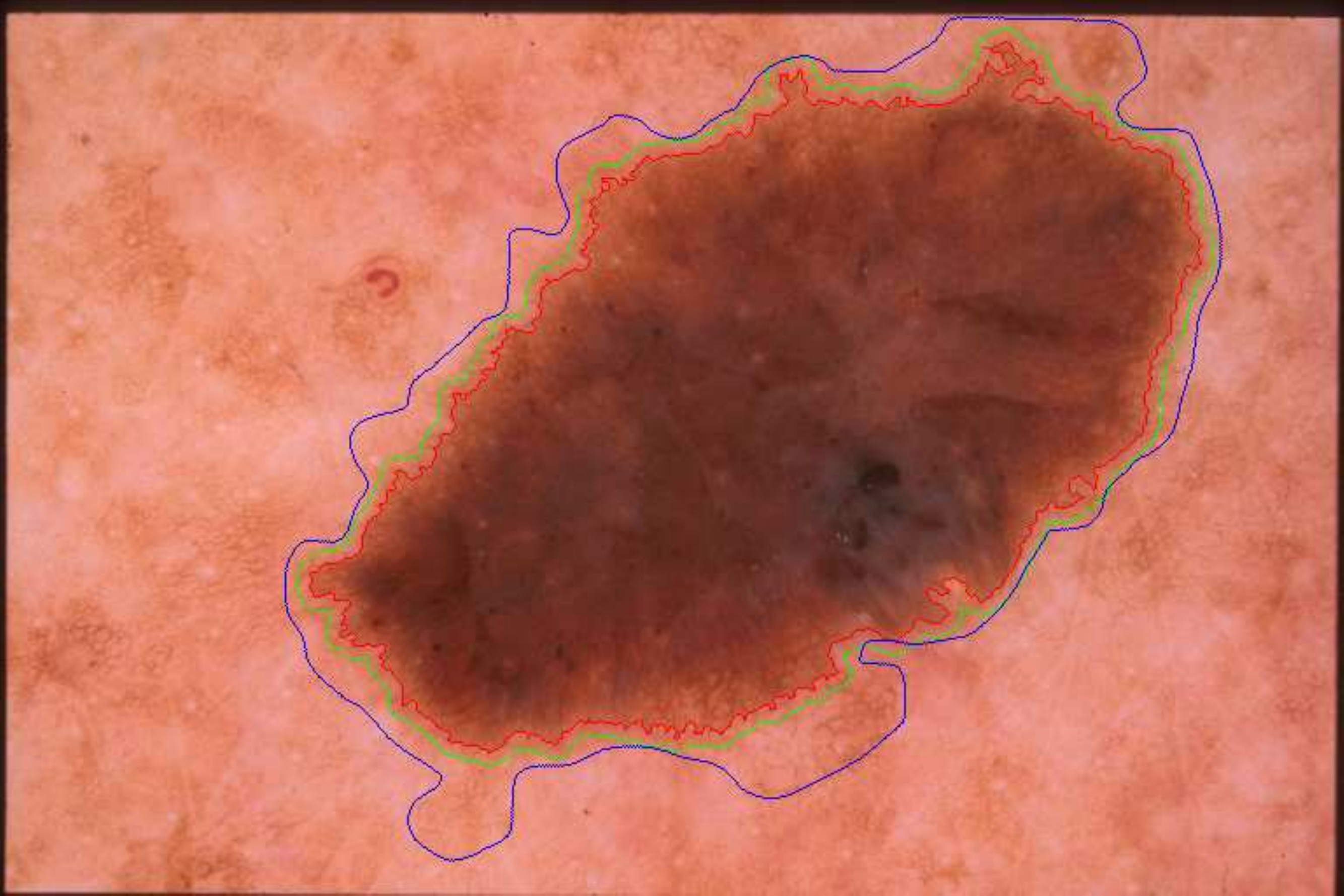}}
 \hspace{.05in}
 \subfigure[Benign ($\varepsilon = 16.25\%$)]{\label{result_h}\includegraphics[width=0.36\columnwidth]{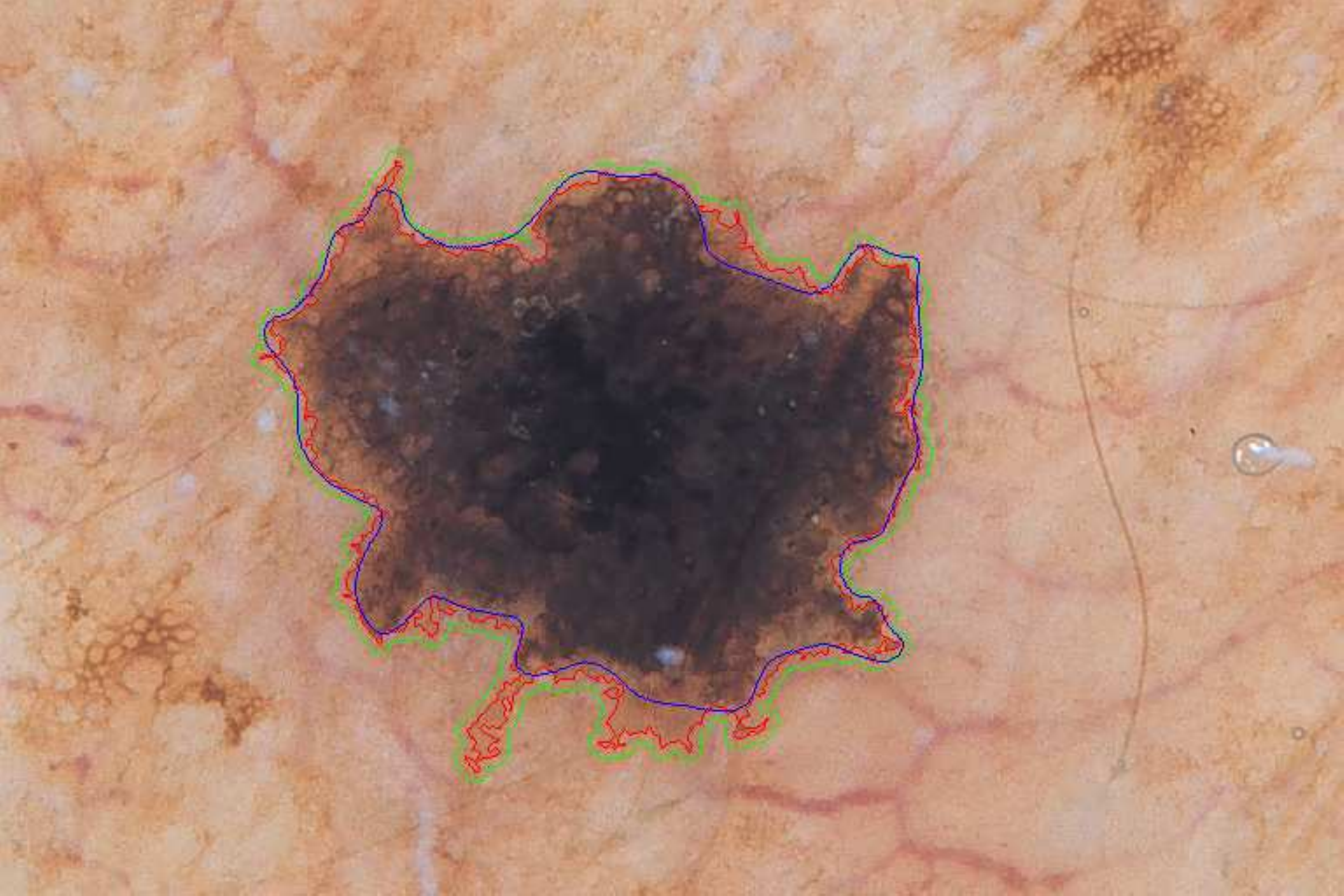}}
 \caption{Sample border detection results}
 \label{fig_results}
\end{figure}

Fig.\ \ref{fig_results} shows sample border detection results obtained by the proposed method. It can be seen that the method performs well even in the presence of complicating factors such as diffuse edges, blood vessels, and skin lines.

\section{Conclusions}
\label{sec_conc}
In this paper, an automated threshold fusion method for detecting lesion borders in dermoscopy images was presented. Experiments on a difficult set of images demonstrated that this method compares favorably to nine recent border detection methods. In addition, the method is easy to implement, extremely fast (0.1 seconds for a typical image of size $768 \times 512$ pixels on an Intel QX9300 2.53GHz computer) and it does not require sophisticated postprocessing.
\par
The presented method may not perform well on images with significant amount of hair or bubbles since these elements alter the histogram, which in turn results in biased threshold computations. For images with hair, a preprocessor such as \emph{DullRazor\texttrademark} \cite{Lee97} might be helpful. Unfortunately, the development of a reliable bubble removal method remains an open problem.

\section*{Acknowledgments}
This publication was made possible by grants from the Louisiana Board of Regents (LEQSF2008-11-RD-A-12), US National Science Foundation (0959583, 1117457), and National Natural Science Foundation of China (61050110449, 61073120).

\end{document}